\documentclass[runningheads]{llncs}

\usepackage{accv}

\usepackage{accvabbrv}

\usepackage{graphicx}
\usepackage{booktabs}
\usepackage{diagbox}
\usepackage{float}
\usepackage{makecell}
\usepackage{subfloat}

\usepackage[accsupp]{axessibility}  

\usepackage{hyperref}

\usepackage{orcidlink}

\DeclareMathOperator*{\argmax}{\arg\!\max}

\begin{document}
\title{JPEG~AI Image Compression Visual Artifacts: Detection Methods and Dataset} 
 \titlerunning{JPEG AI Artifacts: Detection and Dataset}

\author{Daria Tsereh\inst{1,2}, Mark Mirgaleev\inst{2}, Ivan Molodetskikh\inst{1}, Roman Kazantsev\inst{1}, Dmitriy Vatolin\inst{1,2}}

\authorrunning{D.~Tsereh et al.}

\institute{MSU Institute for Artificial Intelligence \and Lomonosov Moscow State University \\
\email{\{daria.romanova, mark.mirgaleev, ivan.molodetskikh, roman.kazantsev, dmitriy\}@graphics.cs.msu.ru}}

\maketitle

\begin{abstract}
Learning-based image compression methods have improved in recent years and started to outperform traditional codecs. However, neural-network approaches can unexpectedly introduce visual artifacts in some images. We therefore propose methods to separately detect three types of artifacts (texture and boundary degradation, color change, and text corruption), to localize the affected regions, and to quantify the artifact strength. We consider only those regions that exhibit distortion due solely to the neural compression but that a traditional codec recovers successfully at a comparable bitrate. We employed our methods to collect artifacts for the JPEG AI verification model with respect to HM-18.0, the H.265 reference software. We processed about 350,000 unique images from the Open Images dataset using different compression-quality parameters; the result is a dataset of 46,440 artifacts validated through crowd-sourced subjective assessment. Our proposed dataset and methods are valuable for testing neural-network-based image codecs, identifying bugs in these codecs, and enhancing their performance. We make source code of the methods and the dataset publicly available.
\end{abstract}

\section{Introduction}

In 2017, Ballé published work that became fundamental to developing end-to-end learning-based methods for image compression~\cite{balle2017endtoendoptimizedimagecompression}.
Its main ideas were Gaussianization of image densities using generalized divisive normalization, a proof of
equivalence for the rate-distortion problem and variational autoencoder, and differentiable approximation of quantization for end-to-end optimization.
Since then, researchers have applied deep learning to image compression and demonstrated an advantage over traditional methods.
More than 500 papers on neural image compression were published in 2023 alone~\cite{10091784,10018275,Liu_2023_CVPR}.

Neural compression's potential has drawn attention from both the academy and industry.
This interest led to the 2022 initiation of the JPEG~AI standardization effort, which seeks to develop the first neural-based image compression standard~\cite{ascenso2023jpegai}.
The JPEG~AI verification model has already demonstrated more than a 10\% BD-rate (PSNR) improvement relative the classic VVC intra codec~\cite{jia2024bitdistribution}.

A learning-based compression pipeline contains artificially designed blocks that convert a complex input signal
into  a simpler form, such as a Gaussian distribution, for more efficient entropy coding.
Unlike traditional codecs, all coding blocks are differentiable and the neural codec can be optimized end to end so they align efficiently.
But because neural coding blocks are complex in their interpretability and predictability for the full range of images, they can produce unexpected results in the form
of visual artifacts. A traditional compression pipeline includes simpler human-interpretable coding blocks from which we can expect concrete artifacts
such as blocking, ringing, and blurring. The artifacts and images subject to them when using neural codecs, however, are undetermined.
We therefore chose to study neural-compression artifacts, develop metrics that detect these artifacts, and create a dataset of such images.
The dataset allows testing and debugging of neural methods to enable improvements.


Our project's objective was to develop metrics sensitive to neural artifacts, such as distortion of text, color, textures, and borders, in compressed images. We designed these metrics to detect even minor distortions that degrade image perception. In this context, an artifact is a distortion, relative to the original image, that is present in an image compressed by a neural codec but it is absent or less noticeable in an image compressed by a classical codec. Our metrics identify images that are less resilient to neural compression techniques compared with traditional techniques.

Using these metrics, we compiled a dataset containing 46,440 images with various artifact types and compression ratios. Additionally, we performed a subjective verification of the automatically detected artifacts to ensure accuracy.

Our main contributions are the following:
\begin{enumerate}
    \item Detection methods for three types of neural-network-compression artifacts that perform better compared to existing methods.
    \item Subjective comparisons to validate identified artifacts.
    \item A dataset containing 46,440 examples of JPEG~AI compression artifacts.
\end{enumerate}

\section{Related Work}

Recent efforts have sought to develop the JPEG~AI standard~\cite{ascenso2023jpegai}. Their results have emphasized the image-compression potential of neural methods, which have become central to further research and development in this area. But JPEG, accounting for more than 800 published papers on neural-based image compression, is not the only group researching this topic. Recent works~\cite{ladune2023coolchic, he2022elic, Liu_2023_CVPR} even outperform the VVC standard~\cite{vvcsoftware}, a top classical image- and video-compression standard, both in terms of PSNR and MS-SSIM~\cite{wang2003multiscale}. Works that propose improvements to the JPEG~AI standard~\cite{jia2024bitdistribution, jia2024bitrate, pan2023lowcomplexity} are also appearing.

\subsection{Image-Quality-Assessment Methods}
Thus far, there had been little research on algorithms that specifically target quality assessment of images compressed by neural codecs.

Many objective metrics handle image-quality assessment. For example, the JPEG~AI standard employs PSNR, MS-SSIM~\cite{wang2003multiscale}, IW-SSIM~\cite{wang2011information}, VIF~\cite{sheikh2004image}, FSIM~\cite {zhang2011fsim}, NLPD~\cite{laparra2016perceptual}, and VMAF~\cite{li2019toward} to evaluate compression models.

In~\cite{isoiec2019}, the developers of the JPEG~AI standard showed that these objective metrics are worse at assessing the quality of learning-based image compression than they are at assessing traditional compression because the correlation with subjective assessments is lower on average for neural codecs than for traditional ones. For instance, even the metric with the best correlation, MS-SSIM~\cite{wang2003multiscale}, performed 7\% worse on neural compression than on classical compression.

This result suggests that existing image-quality-evaluation methods fail to take into account the specifies. They also show low sensitivity to small-area artifacts, which can greatly affect human image perception. 


\subsection{Artifact Detection Methods}
Several studies consider algorithms for evaluating classical-compression artifacts, such as blockiness~\cite{lee2012new} and ringing~\cite{liu2010no-reference,cao2014no-reference}. Moreover, they provide no details about artifact location.

Learning-based image-super-resolution methods can also produce images with neural artifacts. Hence, some recent works introduce methods to suppress super-resolution artifacts.
For example, Xie~et~al.~\cite{pmlr-v202-xie23c} propose the DeSRA method for artifact localization and removal. The authors find artifacts by using a reference image from another super-resolution model that is prone to neural artifacts.

Liang~et~al.~\cite{jie2022LDL} introduce a method to patch super-resolution models against texture artifacts. This method computes an artifact map that is then used in the regularization and stabilization during the training process. The authors do not consider other artifact types.

\subsection{Datasets}
Several image datasets are designed for reproducing artifacts that typically occur with classical compression. For example, LIU4K~\cite{liu4k2020benchmark} includes a wide range of images with varying complexity evaluated by no-reference metrics such as BPP, NIQE, BRISQE, and ENIQA. Developers use these images to reproduce classical compression artifacts such as blurring, blockiness, and ringing. Unfortunately, this dataset has limited applicability to our task, as it lacks a detailed partitioning of the artifacts, and is unsuitable for evaluating neural compression methods, whose distortions differ from those of classical alternatives. 

Rylov~et~al.~\cite{rylov2024learning} recently created a dataset for benchmarking learned-based and traditional image codecs. However, the dataset does not specifically consider images that produce neural compression artifacts.

Our proposed neural-artifact dataset contains information about the type, location, and confidence of the artifact in the partitioning, making it more valuable for research in neural-compression-quality assessment.

\section{Proposed Methods}
\label{sec:proposed}

Our attention focuses on artifacts specific to learning-based codecs. To differentiate them from other artifacts that occur during compression, we additionally compressed the source image using a traditional codec and employ the result as a reference image. The idea is to find prominent differences between JPEG~AI and traditional-codec results.

For our analysis and dataset, we chose intra-frame coding from the HM-18.0 codec~\cite{hevc} as a representative traditional image-compression algorithm that is both widely used and fast enough for batch processing. The target bitrates approximate the JPEG~AI's compression rate. Section~\ref{sec:dataset-overview} describes this process in more detail. For visual comparison we show results from VTM-20.0~\cite{vvcsoftware} with the bitrate chosen to approximate that of JPEG~AI.

All our proposed methods for detecting artifacts take three images as input: an original image \(I_{\text{orig}}\), the image compressed by the neural codec \(I_{\text{neural}}\) (in our case, JPEG~AI), and the image compressed by the traditional codec \(I_{\text{trad}}\). The output is a bounding box of the artifact and a confidence value.

\subsection{Texture-Artifact Detection}
\label{sec:texture-1}

To focus on the image's textured regions, we compute a mask based on the pixel-wise spatial-information (SI) metric~\cite{ITU-T1999}:
\begin{equation}
    M_{x,y} =
    \begin{cases}
        0, & \text{SI}\left( I_{\text{orig}} \right)_{x,y} < 0.05,\\
        1, & \text{otherwise}.
    \end{cases}
\end{equation}

We locate artifacts in the compressed images by computing MS-SSIM~\cite{wang2003multiscale} pixel-wise over the Y channel of the YUV color space. We found that MS-SSIM is more accurate compared to other full-reference metrics for this task.
\begin{align}
    H_{\text{neural}} &= \text{MS-SSIM}(I_{\text{orig}}, I_{\text{neural}}),\\
    H_{\text{trad}} &= \text{MS-SSIM}(I_{\text{orig}}, I_{\text{trad}}).
\end{align}

Next, we subtract these maps and mask them to find textured regions where the traditional codec outperforms JPEG~AI. We also apply average pooling of size 128 to smooth out the outliers:
\begin{equation}
    \Delta H = \text{AvgPool}_{128}\left( \left(H_{\text{trad}} - H_{\text{neural}}\right) \odot M \right).
\end{equation}

The coordinates of the highest value in the resulting map indicate the center point of the artifact, while the value itself is the method's confidence:
\begin{align}
    \text{Center} &= \argmax_{x,y}\left(  \Delta H \right),\\
    \text{Confidence} &= \max_{x,y}\left( \Delta H \right).
\end{align}

In case of multiple centers with a global maximum confidence value, we choose the first occurrence.

\begin{figure}[t]
\centering
\includegraphics[width=1\textwidth]{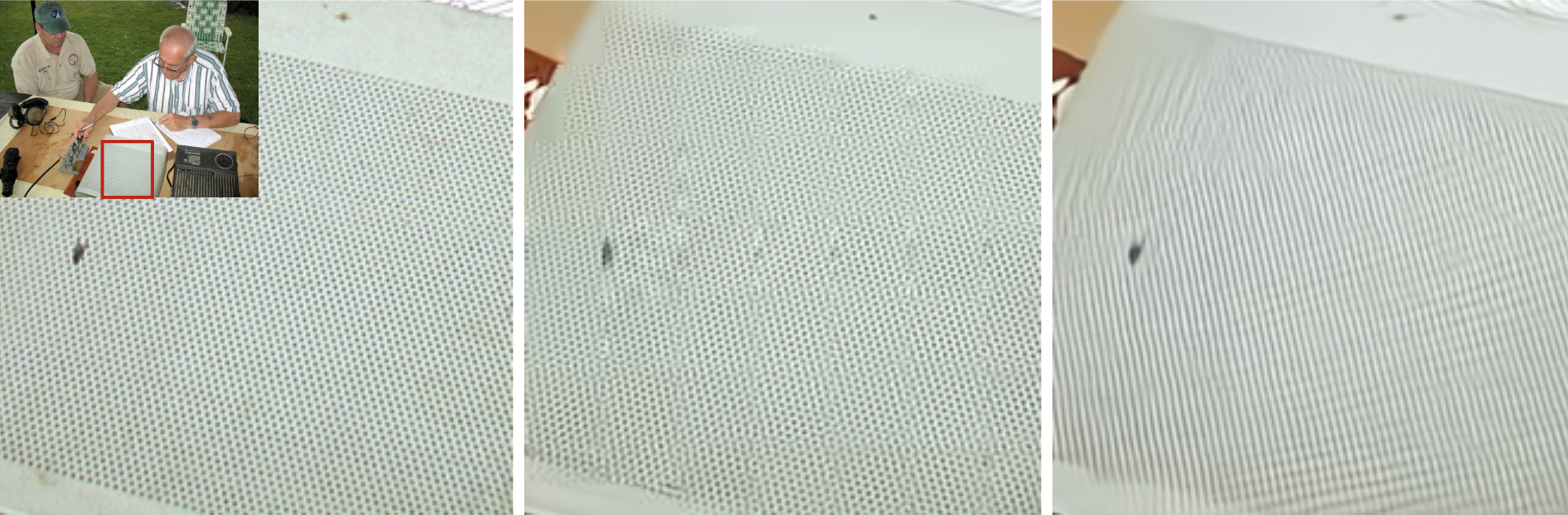}
\begin{minipage}[t]{0.33\textwidth}
    \centering
    Original
\end{minipage}%
\begin{minipage}[t]{0.33\textwidth}
    \centering
    VTM-20.0, 165~times~smaller
\end{minipage}%
\begin{minipage}[t]{0.33\textwidth}
    \centering
    JPEG~AI~4.6, tools~on, high, 170~times~smaller
\end{minipage}
\caption{Texture artifact found by method~\ref{sec:texture-1}. JPEG~AI replaces most of the holes with lines going in different directions, whereas VTM correctly restores most of these holes. Compression ratio is relative to the original image.}
\label{fig:texture1}
\end{figure}

\begin{figure}[t]
\centering
\includegraphics[width=1\textwidth]{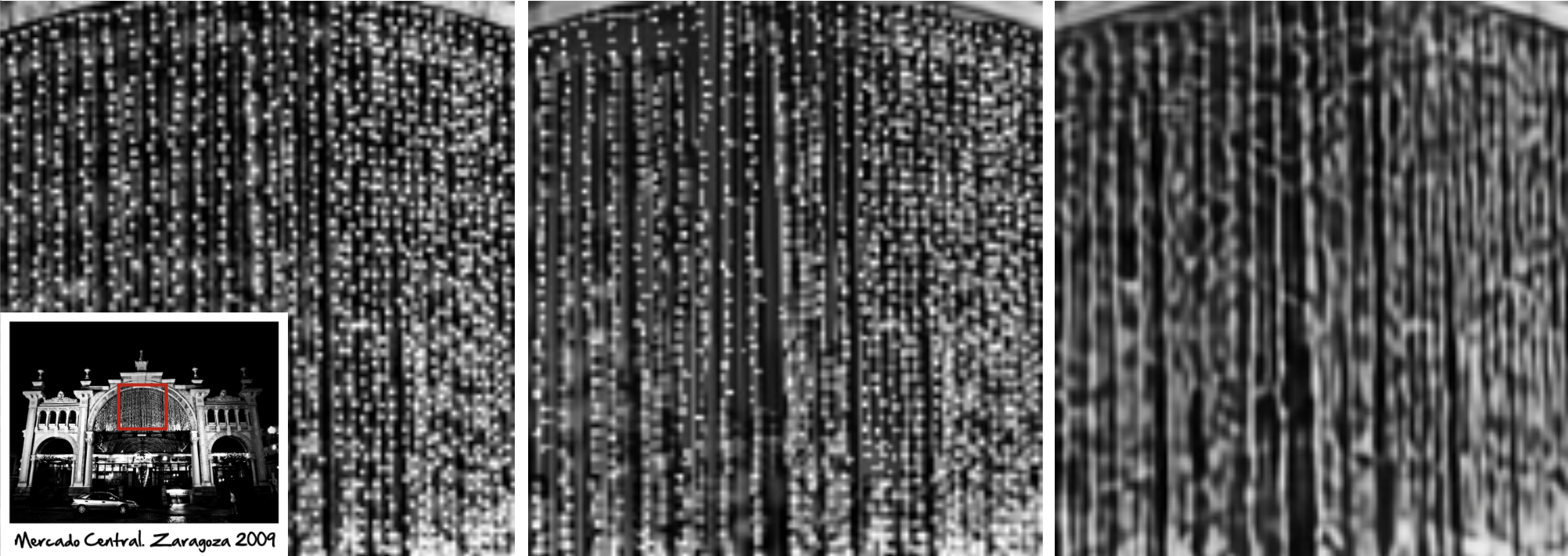}
\begin{minipage}[t]{0.33\textwidth}
\centering
Original
\end{minipage}%
\begin{minipage}[t]{0.33\textwidth}
\centering
VTM-20.0, 208~times~smaller
\end{minipage}%
\begin{minipage}[t]{0.33\textwidth}
\centering
JPEG~AI~4.6, tools~on, high, 204~times~smaller
\end{minipage}
\caption{Texture artifact found by method~\ref{sec:texture-1}. JPEG~AI blurs the texture considerably and reduces detail compared with VTM. Compression ratio is relative to the original image.}
\label{fig:texture2}
\end{figure}

Figures~\ref{fig:texture1} and~\ref{fig:texture2} show texture distortion examples located using this method.

\subsection{Boundary-Texture-Distortion Detection}
\label{sec:texture-2}

Our method for detecting boundary-texture distortions identifies image regions where boundary reconstruction in the neural-compressed image is worse than that in the traditionally compressed one.

First, we locate boundaries in the original image using the Canny~\cite{canny1986computational} detector.
\begin{equation}
B_{x,y} =
\begin{cases} 
1, & \text{if pixel (x, y) belongs to a boundary}, \\
0, & \text{otherwise}.
\end{cases}
\end{equation}

The next step is to compute gradients for the Y channel of all three input images---denoted by \( G_{\text{orig}} \), \( G_{\text{neural}} \), and \( G_{\text{trad}} \)---using the Sobel operator. We compute pixel-wise cosine distances between the gradients as follows:
\begin{align}
H_{\text{neural}} &= \cos \angle(G_{\text{orig}}, G_{\text{neural}}),\\
H_{\text{trad}} &= \cos \angle(G_{\text{orig}}, G_{\text{trad}}).
\end{align}

We take the difference of these maps, mask it with the boundary map, and apply average pooling with size 32 and step 16:
\begin{equation}
\Delta H = \text{AvgPool}_{32}\left(\left(H_{\text{trad}} - H_{\text{neural}}\right) \odot B\right).
\end{equation}

Similarly to the previous method, we locate the highest value in the resulting map and use it as the center point of the artifact, while the value itself is the method's confidence:
\begin{align}
    \text{Center} &= \argmax_{x,y}\left(  \Delta H \right),\\
    \text{Confidence} &= \max_{x,y}\left( \Delta H \right).
\end{align}

\begin{figure}[t]
    \centering
    \includegraphics[width=1\textwidth]{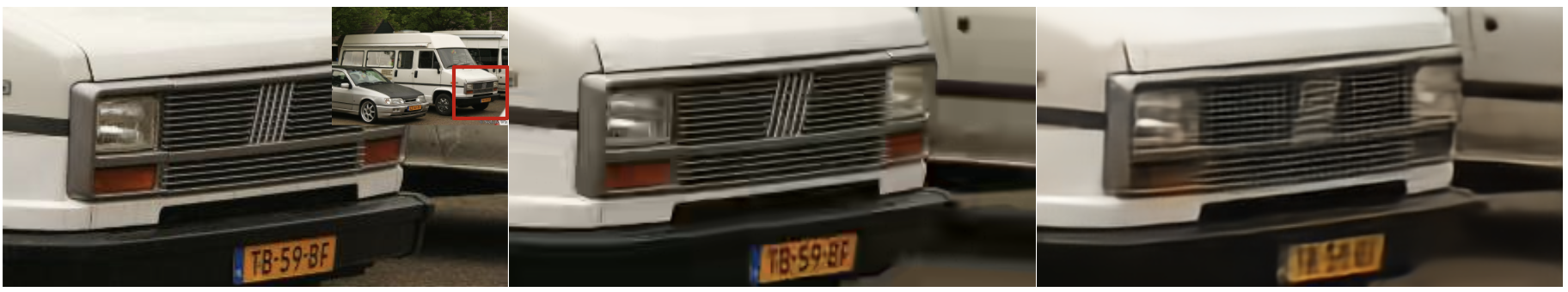}
    \begin{minipage}[t]{0.33\textwidth}
        \centering
        Original
    \end{minipage}%
    \begin{minipage}[t]{0.33\textwidth}
        \centering
        VTM-20.0, intra, 237~times~smaller
    \end{minipage}%
    \begin{minipage}[t]{0.33\textwidth}
        \centering
        JPEG~AI~4.6, tools on, high, 238~times~smaller
    \end{minipage}
    \caption{Boundary artifact found by method~\ref{sec:texture-2}. The edges of the car grill in the JPEG~AI-compressed image changed direction compared with the original, whereas VTM avoids this distortion. Compression ratio is relative to the original image.} 
    \label{fig:edges}
\end{figure}

Figure~\ref{fig:edges} shows a boundary-distortion example located using this method.

\subsection{Large-Color-Distortion Detection}
\label{sec:color-1}

To find large color distortion regions, we measure color differences with the widely used CIEDE2000~\cite{sharma2005ciede2000} metric:
\begin{align}
    H_{\text{neural}} &= \text{CIEDE2000}(I_{\text{orig}}, I_{\text{neural}}),\\
    H_{\text{trad}} &= \text{CIEDE2000}(I_{\text{orig}}, I_{\text{trad}}).
\end{align}

We then filter outliers by zeroing out values less than 3 or greater than 8, denoted by $H'_{\text{neural}}$ and $H'_{\text{trad}}$. Next is an average pooling operation with size 128 and step 64. We take the difference between the resulting maps and locate the highest value, as in the previous methods:
\begin{align}
    \Delta H = \text{AvgPool}_{128}&\left(H'_{\text{trad}}\right) - \text{AvgPool}_{128}\left(H'_{\text{neural}}\right),\\
    \text{Center} &= \argmax_{x,y}\left(  \Delta H \right),\\
    \text{Confidence} &= \max_{x,y}\left( \Delta H \right).
\end{align}

\begin{figure}[t]
    \centering
    \includegraphics[width=1\textwidth]{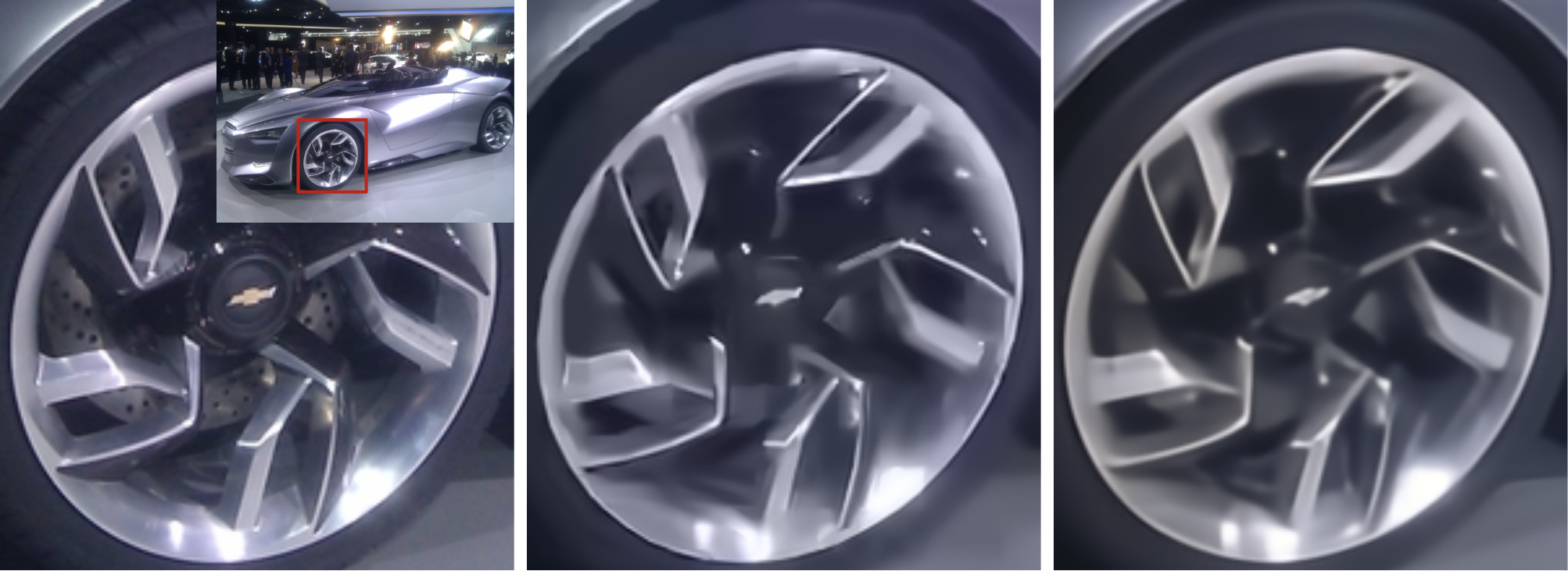}
    \begin{minipage}[t]{0.33\textwidth}
        \centering
       Original
    \end{minipage}%
    \begin{minipage}[t]{0.33\textwidth}
        \centering
       VTM-20.0, 305~times~smaller
    \end{minipage}%
    \begin{minipage}[t]{0.33\textwidth}
        \centering
       JPEG~AI~4.6, tools~on, high, 303~times~smaller
    \end{minipage}
    \caption{Color artifact found by method~\ref{sec:color-1}. The wheel's hue differs between the JPEG~AI result and the original image, whereas VTM restores it correctly. Compression ratio is relative to the original image.}
    \label{fig:color1}
\end{figure}

Figure~\ref{fig:color1} shows a color-distortion example located using this method.

\subsection{Small-Color-Artifact Detection}
\label{sec:color-2}

In contrast to the previous method, small-color-artifact detection finds smaller-scale color distortions. We mostly follow the super-resolution-artifact detection method in~\cite{jie2022LDL}, adapting it to compression.

Our approach follows the residual- and primary-map-construction steps from~\cite{jie2022LDL} for both neural- and traditional-codec results, denoted by $r \in \left\{\text{neural},\text{trad}\right\}$:
\begin{align}
    R^C_r(i, j) &= \sum_{c \in C}\left(\left|I^c_{\text{orig}}(i, j) - I^c_{r}(i, j)\right|\right),\\
    M^C_r(i, j) &= var(R^C_r(i - \frac{n - 1}{2}:i + \frac{n - 1}{2}, j - \frac{n - 1}{2}:j + \frac{n - 1}{2})),
\end{align}
where $C$ indicates the color channels on which we will compute these maps. We set $n$ to 33 empirically.

\begin{figure}[t]
    \centering
    \includegraphics[width=1\textwidth]{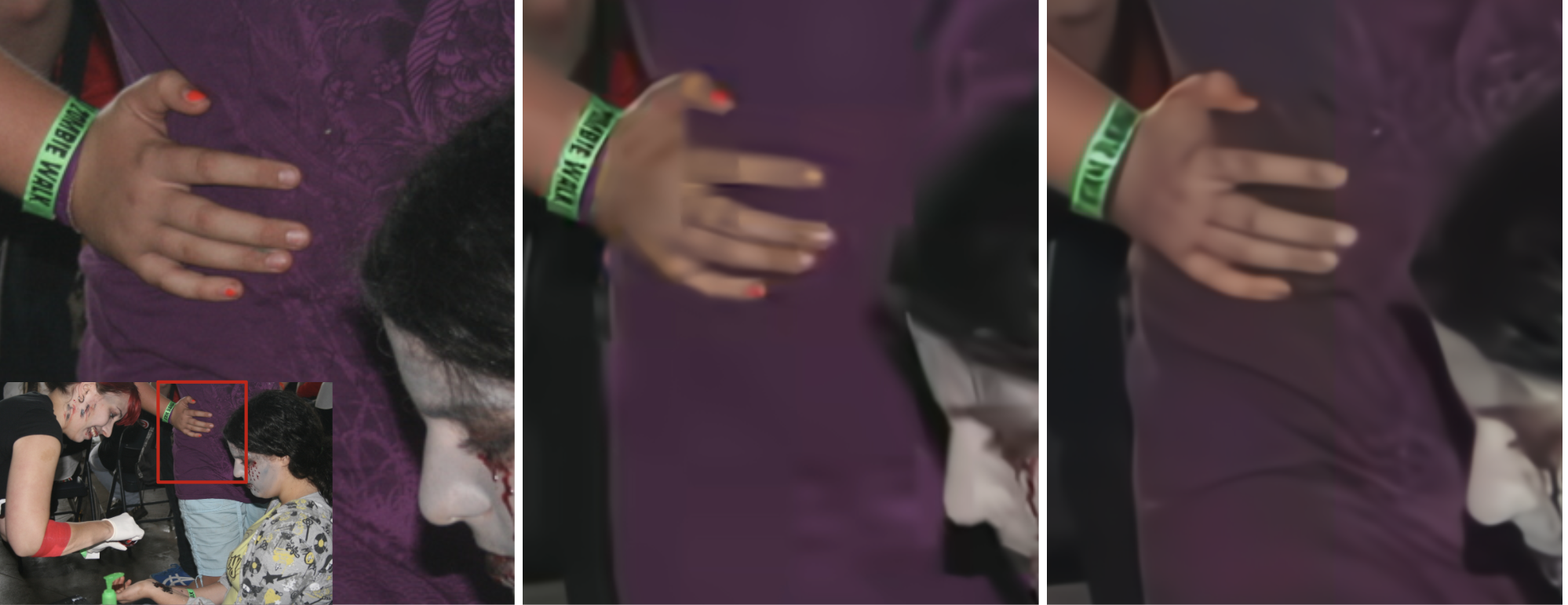}
    \begin{minipage}[t]{0.33\textwidth}
        \centering
       Original
    \end{minipage}%
    \begin{minipage}[t]{0.33\textwidth}
        \centering
        VTM-20.0, 311~times~smaller
    \end{minipage}%
    \begin{minipage}[t]{0.33\textwidth}
        \centering
       JPEG~AI~4.6, tools~on, high, 310~times~smaller
    \end{minipage}
    \caption{Color artifact found by method~\ref{sec:color-2}. JPEG~AI failed to restore the red nail color and darkened the cloth, whereas VTM restored them faithfully. Compression ratio is relative to the original image.}
    \label{fig:color2}
\end{figure}

We also compute the stable patch-level variance and use it to scale the primary map:
\begin{align}
    \sigma^C_r &= (var(R^C_r))^\frac15,\\
    S^C_r(i, j) &= \sigma^C_r M^C_r(i, j).
\end{align}

Next we take the difference between the resulting maps for neural- and traditional-compression results, then apply a threshold:
\begin{align}
    \Delta S^C &= S^C_\text{neural} - S^C_\text{trad},\\
    B^C &=
    \begin{cases}
        1, & \Delta S^C > T,\\
        0, & \text{otherwise},
    \end{cases}
\end{align}
with $T$ empirically selected to be 0.0015.

We locate the artifacts by performing these computations on the U and~V channels of the images' YUV color representation as well as on the a and~b channels of their Lab color representation, then multiply the resulting binary maps:
\begin{align}
    B = B^{U,V} * B^{a,b}.
\end{align}

The final step is to locate distinct contours on the resulting binary map $B$ using algorithm proposed in~\cite{suzuki1985topological} as implemented in OpenCV~\cite{opencv_library} and designate their bounding rectangles as the artifact locations. The confidence is the maximum of the scaled primary maps inside the respective bounding rectangle:
\begin{equation}
    \text{Confidence} = \max(\Delta S^{U,V}, \Delta S^{a, b}).
\end{equation}

Figure~\ref{fig:color2} shows a color-artifact example located using this method.

\subsection{Text-Artifact Detection}
\label{sec:text-1}

\begin{figure}[t]
    \centering
    \begin{minipage}[b]{0.75\textwidth}
        \centering
        \includegraphics[width=\textwidth]{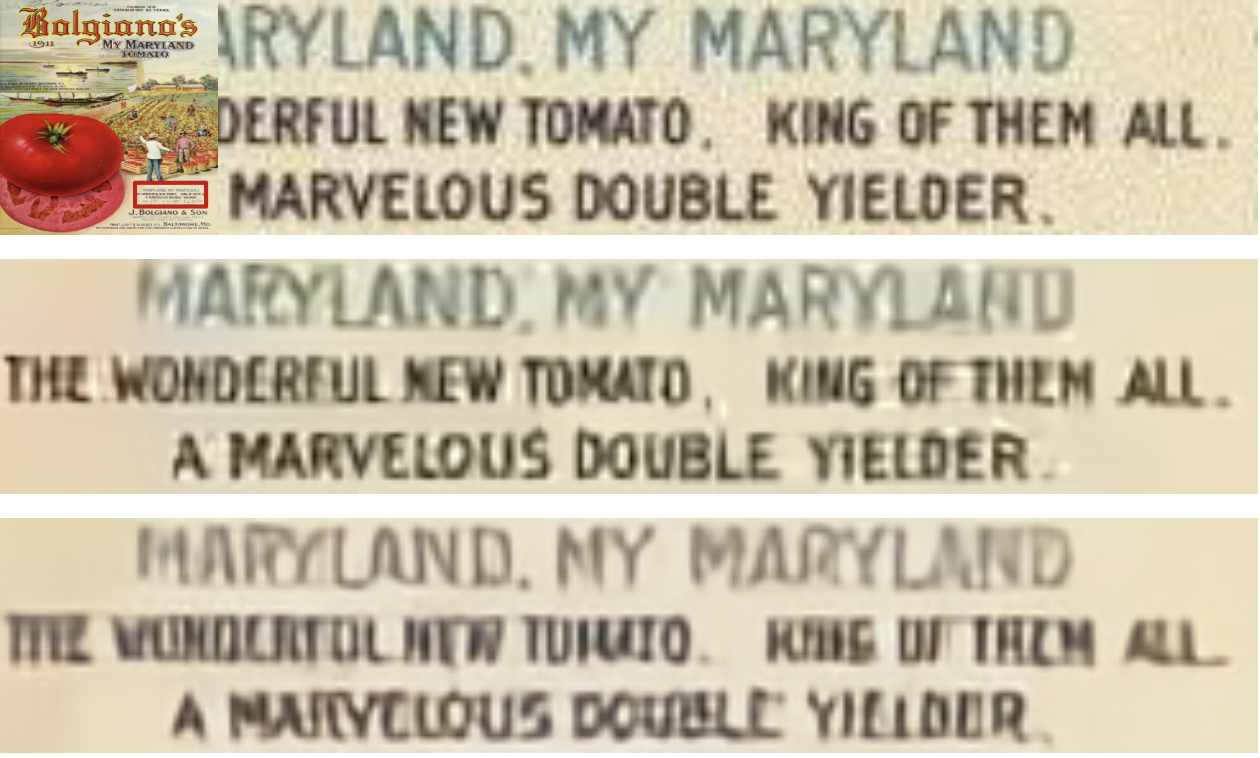}
    \end{minipage}%
    \begin{minipage}[b]{0.25\textwidth}
        \centering
        Original\\
        \vspace{37pt}
        VTM-20.0,\\
        172~times~smaller \\
        \vspace{25pt}
        JPEG~AI~4.6, \\
        tools~on, high, \\
        170~times~smaller
        \vspace{11pt}
    \end{minipage}
    \caption{Text artifact found by method~\ref{sec:text-1}. The readability of the text in the JPEG~AI result is worse than that in the VTM result. Compression ratio is relative to the original image.}
    \label{fig:text1}
\end{figure}

Our approach to detecting text artifacts begins by running the OpenMMLab text detector~\cite{mmocr} on $I_\text{orig}$, returning a list of polygons that encompass any text along a set of with confidence scores. Polygons with confidence less than 0.7 are discarded to reduce false positives.

Next, we surround each polygon with a bounding box and discard bounding boxes whose area is less than 20×20 pixels. Distortions in such small areas have proven to exhibit minimal impact on image perception.

Our method processes each remaining bounding box individually. We crop the area corresponding to the bounding box from each of the three input images to obtain $C_\text{orig}, C_\text{neural}, C_\text{trad}$. We compute the  feature-similarity index (FSIM)~\cite{zhang2011fsim} between the original and compressed images. Higher FSIM values indicate a greater similarity between crops. The confidence value for the bounding box is the difference between the FSIM values:
\begin{align}
    H_\text{neural} &= \text{FSIM}(C_\text{orig}, C_\text{neural}),\\
    H_\text{trad} &= \text{FSIM}(C_\text{orig}, C_\text{trad}),\\
    \text{Confidence} &= H_\text{trad} - H_\text{neural}.
\end{align}
For crop-similarity computation we experimented with other visual quality metrics such as PSNR, MS-SSIM, and DIST. FSIM-based text artifact classifier showed better results.

Next, we enlarge all bounding boxes to 300×300 pixels for subsequent visual analysis and merge those with significant overlap (intersection-over-union > 0.12). The confidence value for the merged bounding box is the maximum of the values for the original boxes.

\begin{figure}[t]
    \centering
    \includegraphics[width=1\textwidth]{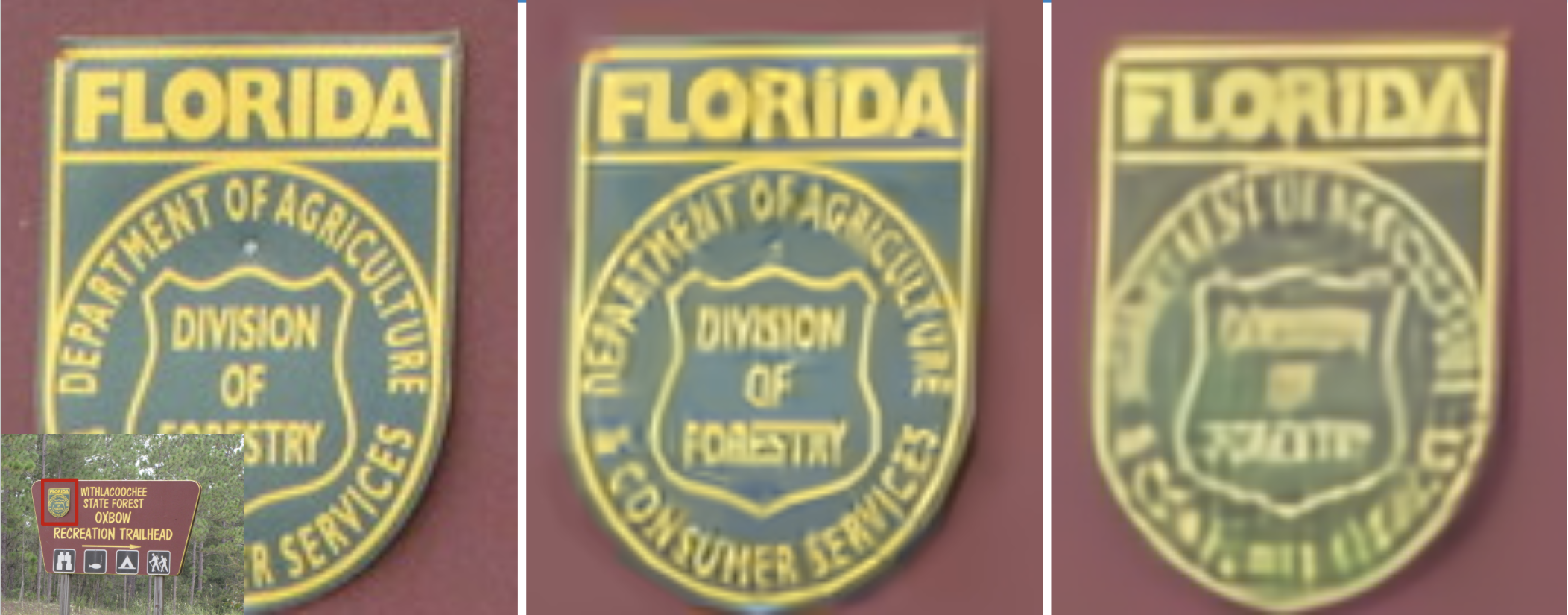}
    \begin{minipage}[t]{0.33\textwidth}
        \centering
        Original
    \end{minipage}%
    \begin{minipage}[t]{0.33\textwidth}
        \centering
        VTM-20.0, \\176~times~smaller
    \end{minipage}%
    \begin{minipage}[t]{0.33\textwidth}
        \centering
        JPEG~AI~4.6, tools~on, high, 175~times~smaller
    \end{minipage}
    \caption{Text artifact found by method~\ref{sec:text-1}. The readability of the text in the JPEG~AI result is worse than that in the VTM result. Compression ratio is relative to the original image.}
    \label{fig:text2}
\end{figure}

Figures~\ref{fig:text1} and~\ref{fig:text2} show text-distortion examples located using this method.

\section {JPEG~AI-Compression-Artifact Dataset}
\label{sec:dataset}

To facilitate JPEG~AI and other learning-based-codec research, we collected a dataset of 46,440 JPEG~AI-artifacts of three types (texture, color, and text distortions). The dataset includes the source images and detailed annotations for every artifact, including its type and bounding box. We validated all examples through crowdsourced subjective assessment. This section describes our procedures for dataset collection, annotation, and validation.

\subsection{Overview}
\label{sec:dataset-overview}

We collected 350,000 source images from the Open Images~\cite{kuznetsova2020open} dataset, compressed them using JPEG~AI and a traditional codec, and processed them using the methods described in Section~\ref{sec:proposed} to obtain the preliminary annotations. Our next step was to conduct a subjective assessment, described in Section~\ref{sec:subjective}, to verify the annotations and discard false positives. In total, we obtained 46,440 confirmed artifacts.

As of this writing, JPEG~AI remains under development, and new versions emerge regularly. We followed the development process and switched to new JPEG~AI versions as they arrived. Our dataset therefore contains examples from several versions. We set the tools parameter to off for all JPEG~AI versions. Our example figures demonstrate that the artifacts remain on the tools on configuration.

JPEG~AI offers five quality presets that control its quality/size tradeoff. We compressed and processed images using all five quality presets. After subjective validation, we deduplicated the artifacts on the basis of intersection-over-union, preferring those that occurred at higher quality presets because confirmed artifacts at these presets tend to be scarcer and are therefore more valuable.

Table~\ref{tab:dataset-stats} summarizes the partitions of our dataset by artifact type, by the JPEG~AI quality preset, and by the JPEG~AI version used to compress the source image.

For the traditional-codec-compression results we used intra coding from VTM-20.0 and HM-18.0, the reference implementations of the VVC and HEVC standards, respectively. We made the target VTM bitrate as close as possible to the file size produced by JPEG~AI for each of its quality presets. Then, since HM uses a previous-generation coding standard, we raised its bitrate by 20\% compared with VTM to achieve a similar visual quality. The next step employed HM-compressed images as the \(I_{\text{trad}}\) input to our detection methods. We were unable to use the VTM results directly because that method's compression speed makes collecting a large-scale dataset impractical.

\begin{table}[t]
    \centering
    \begin{minipage}[t]{.31\textwidth}%
        \vspace{0pt}
        \begin{tabular}{c|c}
            \textbf{\makecell{Artifact \\ Type}} & \textbf{\# Artifacts} \\
            \hline
            Texture & 11,932 \\
            Color & 7,545 \\
            Text & 26,963 \\
        \end{tabular}
    \end{minipage}%
    \begin{minipage}[t]{.35\textwidth}%
        \vspace{0pt}
        \begin{tabular}{c|c}
            \textbf{\makecell{Quality \\ Parameter}} & \textbf{\# Artifacts} \\
            \hline
            006 & 27,784 \\
            012 & 10,350 \\
            025 & 6,916 \\
            050 & 200 \\
            075 & 1,190 \\
        \end{tabular}
    \end{minipage}%
    \begin{minipage}[t]{.31\textwidth}%
        \vspace{0pt}
        \begin{tabular}{c|c}
            \textbf{\makecell{Version, \\ Preset}} & \textbf{\# Artifacts} \\
            \hline
            3.3, base & 4,162 \\
            4.1, base & 37,035 \\
            4.1, high & 4,083 \\
            4.2, base & 1,160 \\
        \end{tabular}
    \end{minipage}%
    \vspace{3pt}
    \caption{Statistics for our dataset. From left to right: artifact count by type, by JPEG~AI quality parameter, by JPEG~AI version and preset.}
    \label{tab:dataset-stats}
\end{table}

\subsection{Subjective Validation}
\label{sec:subjective}

We selected the Toloka\footnote{\url{https://toloka.ai}} platform for crowdsourced comparisons. Each artifact type underwent validation using a separate labeling task, accompanied by instructions to familiarize participants with images that contain artifacts. Every task had control examples to ensure accurate responses. Each artifact was judged by three participants, and one positive vote was considered sufficient to confirm the artifact.

For texture-artifact validation, participants were asked to evaluate 300×300-pixel crops from $I_\text{orig}, I_\text{neural}$, and $I_\text{trad}$, centered on the artifact located by methods~\ref{sec:texture-1} and~\ref{sec:texture-2}. They were asked which contained the greater texture distortion: the crop from JPEG~AI or the one from HM. If they chose JPEG~AI, the artifact was considered confirmed.

For color-artifact validation, we similarly cropped 300×300-pixel fragments and combined crops from $I_\text{orig}$ and $I_\text{trad}$, as well as from $I_\text{orig}$ and $I_\text{neural}$. The result was two checkerboard images that made the color differences more apparent. Participants were then asked to select the image in which the checkerboard was most prominent---i.e., the one in which the color distortion was more pronounced.

For text-artifact validation, participants were asked to rate pairs of crops from $I_\text{neural}$ and $I_\text{trad}$ found by method~\ref{sec:text-1}. These comparisons omitted the original crop because the original text is unnecessary to confirm text distortion. Participants were to choose the image in which the text was more distorted.





\section{Comparison with Existing Methods}

To evaluate the proposed methods' effectiveness, we compare them with several existing full-reference image quality assessment methods by area under the ROC curve (AUC). For this comparison, we compiled three balanced test sets, one for each of the artifact types (texture, color, text). Every test set contains 50 confirmed artifacts of the respective type from our dataset, representing the positive cases, and 50 images free of this type of artifact, representing the negative cases. These negative cases are evenly split into completely artifact-free images, and images containing the remaining two types of artifacts. For example, the texture artifact test set contains 50 texture artifacts, 16 artifact-free images, 17 color artifacts and 17 text artifacts. We employ this partitioning to highlight the ability of our methods to distinguish their specific artifact type.

The values for the image quality assessment methods is the difference between their evaluation of the traditionally- and the neural-compressed image:
\begin{align}
\Delta \text{method} &= \text{method}(I_{\text{orig}}, I_{\text{trad}}) - \text{method}(I_{\text{orig}}, I_{\text{neural}})\\
\Delta \text{NLPD} &= \text{NLPD}(I_{\text{orig}}, I_{\text{neural}}) - \text{NLPD}(I_{\text{orig}}, I_{\text{trad}}).
\end{align}
The NLPD equation is flipped because a lower value of this method indicates better image quality, in contrast to other methods.

Table~\ref{tab:auc_all} shows the evaluation results. Each of our proposed methods achieves the highest AUC value for its respective artifact type, confirming its improved ability to detect that specific type of artifact compared to existing methods.

\begin{table}[t]
\centering
\begin{tabular}{r|ccc}
\textbf{Method} & \textbf{\makecell{Texture \\ Artifact Set}} & \textbf{\makecell{Color \\ Artifact Set}} & \textbf{\makecell{Text \\ Artifact Set}} \\
\hline
\textbf{Proposed Texture Method \ref{sec:texture-1}} & \textbf{0.80} & - & - \\
\textbf{Proposed Boundary Method \ref{sec:texture-2}} & \textbf{0.79} & - & - \\
\textbf{Proposed Large-Color Method \ref{sec:color-1}} & - & \textbf{0.83} & - \\
\textbf{Proposed Small-Color Method \ref{sec:color-2}} & - & \textbf{0.63} & - \\
\textbf{Proposed Text Method \ref{sec:text-1}} & - & - & \textbf{0.88} \\
PSNR & 0.67 & 0.55 & 0.61 \\
SSIM & 0.59 & 0.53 & 0.50 \\
MS-SSIM & 0.64 & 0.51 & 0.51 \\
IW-SSIM & 0.64 & 0.41 & 0.57 \\
VIF(P) & 0.71 & 0.54 & 0.69 \\
FSIM & 0.63 & 0.51 & 0.51 \\
NLPD & 0.73 & 0.49 & 0.59 \\
\end{tabular}
\vspace{3pt}
\caption{AUC comparison between proposed and existing methods across different artifact types.}
\label{tab:auc_all} 
\end{table}

\section{Conclusion}

In this paper, we focused on characteristic visual artifacts produced by learning-based compression methods---specifically, JPEG~AI. We proposed automatic methods to detect texture, boundary, color, and text distortions that appear in compression results from a neural codec but not in those from a traditional codec.

Using these methods we collected a dataset of 46,440 JPEG~AI-artifact examples to facilitate further research. All artifacts in the dataset underwent validation through crowdsourced subjective assessment. Experimental evaluation of this dataset showed that our detection methods are better suited to finding learning-based compression artifacts compared with existing methods. Our methods also aid in analyzing future JPEG~AI versions as well as other learning-based codecs.

The dataset and the source code for all proposed methods are publicly available at \url{https://redacted}.

\bibliographystyle{splncs04}
\bibliography{artifacts}

\begin{thebibliography}{10}
\providecommand{\url}[1]{\texttt{#1}}
\providecommand{\urlprefix}{URL }
\providecommand{\doi}[1]{https://doi.org/#1}

\bibitem{mmocr}
{MMOCR: OpenMMLab Toolbox for Text Detection and Recognition}. \url{https://github.com/open-mmlab/mmocr} (2022)

\bibitem{isoiec2019}
Ascenso, J., Akayzi, P., Testolina, M., Boev, A., Alshina, E.: Performance evaluation of learning based image coding solutions and quality metrics. ISO/IEC JTC 1/SC29/WG1 N85013 85th JPEG Meeting, San Jose, USA, November 2019

\bibitem{ascenso2023jpegai}
Ascenso, J., Alshina, E., Ebrahimi, T.: The jpeg ai standard: Providing efficient human and machine visual data consumption. IEEE MultiMedia  \textbf{30}(1),  100--111 (Jan 2023). \doi{10.1109/MMUL.2023.3245919}

\bibitem{balle2017endtoendoptimizedimagecompression}
Ballé, J., Laparra, V., Simoncelli, E.P.: End-to-end optimized image compression (2017), \url{https://arxiv.org/abs/1611.01704}

\bibitem{opencv_library}
Bradski, G.: {The OpenCV Library}. Dr. Dobb's Journal of Software Tools  (2000)

\bibitem{canny1986computational}
Canny, J.: A computational approach to edge detection. IEEE Transactions on Pattern Analysis and Machine Intelligence  (1986)

\bibitem{cao2014no-reference}
Cao, Z., Wei, Z., Zhang, G.: A no-reference sharpness metric based on structured ringing for jpeg2000 images. Advances in Optical Technologies  (2014)

\bibitem{vvcsoftware}
{Fraunhofer Heinrich Hertz Institute}: Versatile video coding reference software version 12.1 (vtm-12.1). \url{https://vcgit.hhi.fraunhofer.de/jvet/VVCSoftware_VTM/-/tags/VTM-12.1} (January 2021)

\bibitem{10018275}
Fu, H., Liang, F., Liang, J., Li, B., Zhang, G., Han, J.: Asymmetric learned image compression with multi-scale residual block, importance scaling, and post-quantization filtering. IEEE Transactions on Circuits and Systems for Video Technology  \textbf{33}(8),  4309--4321 (2023). \doi{10.1109/TCSVT.2023.3237274}

\bibitem{10091784}
Fu, H., Liang, F., Lin, J., Li, B., Akbari, M., Liang, J., Zhang, G., Liu, D., Tu, C., Han, J.: Learned image compression with gaussian-laplacian-logistic mixture model and concatenated residual modules. IEEE Transactions on Image Processing  \textbf{32},  2063--2076 (2023). \doi{10.1109/TIP.2023.3263099}

\bibitem{he2022elic}
He, D., Yang, Z., Peng, W., Ma, R., Qin, H., Wang, Y.: Elic: Efficient learned image compression with unevenly grouped space-channel contextual adaptive coding. In: 2022 IEEE/CVF Conference on Computer Vision and Pattern Recognition (CVPR). pp. 5708--5717. IEEE, New Orleans, LA, USA (2022). \doi{10.1109/CVPR52688.2022.00563}

\bibitem{ITU-T1999}
ITU-T~Recommendation, P.: Subjective video quality assessment methods for multimedia applications  (1999)

\bibitem{hevc}
{ITU-T Video Coding Experts Group and ISO/IEC Moving Picture Experts Group}: {High Efficiency Video Coding (HEVC)}. \url{https://hevc.hhi.fraunhofer.de/} (2013)

\bibitem{jia2024bitrate}
Jia, P., Koyuncu, A.B., Mao, J., Cui, Z., Ma, Y., Guo, T., Solovyev, T., et~al.: Bit rate matching algorithm optimization in jpeg-ai verification model. \url{http://arxiv.org/abs/2402.17487} (2024)

\bibitem{jia2024bitdistribution}
Jia, P., Mao, J., Koyuncu, E., Koyuncu, A.B., Solovyev, T., Karabutov, A., Zhao, Y., Alshina, E., Kaup, A.: Bit distribution study and implementation of spatial quality map in the jpeg-ai standardization. \url{http://arxiv.org/abs/2402.17470} (2024)

\bibitem{kuznetsova2020open}
Kuznetsova, A., Rom, H., Alldrin, N., Uijlings, J., Krasin, I., Pont-Tuset, J., Kamali, S., Popov, S., Malloci, M., Kolesnikov, A., Duerig, T., Ferrari, V.: The open images dataset v4. International Journal of Computer Vision  (2020)

\bibitem{ladune2023coolchic}
Ladune, T., Philippe, P., Henry, F., Clare, G., Leguay, T.: Cool-chic: Coordinate-based low complexity hierarchical image codec. In: 2023 IEEE/CVF International Conference on Computer Vision (ICCV). pp. 13469--13476. IEEE, Paris, France (2023). \doi{10.1109/ICCV51070.2023.01243}

\bibitem{laparra2016perceptual}
Laparra, V., Ball{\'e}, J., Berardino, A., Simoncelli, E.P.: Perceptual image quality assessment using a normalized laplacian pyramid. Electronic Imaging  (2016)

\bibitem{lee2012new}
Lee, S., Park, S.J.: A new image quality assessment method to detect and measure strength of blocking artifacts. Signal Processing: Image Communication  \textbf{27},  31--38 (2012)

\bibitem{li2019toward}
Li, Z., Aaron, A., Katsavounidis, I., Moorthy, A., Manohara, M.: Toward a practical perceptual video quality metric, \url{https://medium.com/netflix-techblog/toward-a-practical-perceptual-video-quality-metric-653f208b9652}

\bibitem{jie2022LDL}
Liang, J., Zeng, H., Zhang, L.: Details or artifacts: A locally discriminative learning approach to realistic image super-resolution. In: Proceedings of the IEEE Conference on Computer Vision and Pattern Recognition (2022)

\bibitem{liu2010no-reference}
Liu, H., Klomp, N., Heynderickx, I.: A no-reference metric for perceived ringing artifacts in images. IEEE Transactions on Circuits and Systems for Video Technology  \textbf{20}(4),  529--539 (April 2010)

\bibitem{liu4k2020benchmark}
Liu, J., Liu, D., Yang, W., Xia, S., Zhang, X., Dai, Y.: A comprehensive benchmark for single image compression artifact reduction. IEEE Transactions on Image Processing  \textbf{29},  7845--7860 (2020)

\bibitem{Liu_2023_CVPR}
Liu, J., Sun, H., Katto, J.: Learned image compression with mixed transformer-cnn architectures. In: Proceedings of the IEEE/CVF Conference on Computer Vision and Pattern Recognition (CVPR). pp. 14388--14397 (June 2023)

\bibitem{pan2023lowcomplexity}
Pan, X., Ding, D., Wang, L., Xu, X., Liu, S.: Low-complexity transform network architecture for jpeg ai image codec. In: 2023 IEEE International Conference on Visual Communications and Image Processing (VCIP). pp.~1--5. IEEE, Jeju, Korea, Republic of (2023). \doi{10.1109/VCIP59821.2023.10402641}

\bibitem{rylov2024learning}
Rylov, V., Kazantsev, R., Vatolin, D.: Learning-based image compression benchmark (2024), \url{https://videoprocessing.ai/benchmarks/learning-based-image-compression.html}

\bibitem{sharma2005ciede2000}
Sharma, G., Wu, W., Dalal, E.N.: The ciede2000 color-difference formula: Implementation notes, supplementary test data, and mathematical observations. Color Research and Application  (2004)

\bibitem{sheikh2004image}
Sheikh, H.R., Bovik, A.C.: Image information and visual quality. In: IEEE International Conference on Acoustics, Speech, and Signal Processing (2004)

\bibitem{suzuki1985topological}
Suzuki, S., et~al.: Topological structural analysis of digitized binary images by border following. Computer vision, graphics, and image processing  \textbf{30}(1),  32--46 (1985)

\bibitem{wang2011information}
Wang, Z., Li, Q.: Information content weighting for perceptual image quality assessment. IEEE Transactions on Image Processing  (2011)

\bibitem{wang2003multiscale}
Wang, Z., Simoncelli, E.P., Bovik, A.C.: Multiscale structural similarity for image quality assessment. In: The Thirty-Seventh Asilomar Conference on Signals, Systems \& Computers (2003)

\bibitem{pmlr-v202-xie23c}
Xie, L., Wang, X., Chen, X., Li, G., Shan, Y., Zhou, J., Dong, C.: {D}e{SRA}: Detect and delete the artifacts of {GAN}-based real-world super-resolution models. In: Proceedings of the 40th International Conference on Machine Learning. Proceedings of Machine Learning Research, vol.~202, pp. 38204--38226. PMLR (23--29 Jul 2023), \url{https://proceedings.mlr.press/v202/xie23c.html}

\bibitem{zhang2011fsim}
Zhang, L., Zhang, L., Mou, X., Zhang, D.: Fsim: A feature similarity index for image quality assessment. IEEE Transactions on Image Processing pp. 2378--2386 (2011). \doi{10.1109/TIP.2011.2109730}

\end{thebibliography}
\end{document}